\documentclass[10pt,twocolumn,letterpaper]{article}

\usepackage[pagenumbers]{cvpr}

\usepackage{cuted} 

\usepackage{currfile} 

\usepackage{caption} 

\usepackage{amsmath,amssymb}
\usepackage{booktabs}
\usepackage{float}
\makeatletter
\renewcommand\paragraph{\@startsection{paragraph}{4}{\z@}%
  {5pt \@plus 1pt \@minus 1pt}%
  {-1em}%
  {\normalfont\normalsize\bfseries}}
\makeatother
\AtEndPreamble{%
  \crefname{equation}{Eq.}{Eqs.}\Crefname{equation}{Eq.}{Eqs.}%
  \crefname{figure}{Fig.}{Figs.}\Crefname{figure}{Figure}{Figures}%
}

\newcommand{\ours}{EquiSD}                 
\newcommand{\ivpsim}{IVP-Sim}

\definecolor{linkblue}{rgb}{0.21,0.49,0.74}
\usepackage[pagebackref,breaklinks,colorlinks,allcolors=linkblue]{hyperref}

\def\paperID{}
\def\confName{}
\def\confYear{}

\title{Teaching Vision-Language Models to Use the Scale They Are Given: Label-Free Equivariance Training for Metric Physical Reasoning}

\author{%
Kaizhen Tan\textsuperscript{1} \qquad
Yang Feng\textsuperscript{2} \qquad
Heqing Du\textsuperscript{2} \qquad
Siru Tao\textsuperscript{3} \qquad
Xin Xu\textsuperscript{3} \qquad
Hanzhe Hong\textsuperscript{3}
\\[2pt]
\normalsize
\textsuperscript{1}New York University \qquad
\textsuperscript{2}Columbia University \qquad
\textsuperscript{3}Carnegie Mellon University
}

\begin{document}
\maketitle
\begin{strip}
  \centering
  \includegraphics[width=\linewidth]{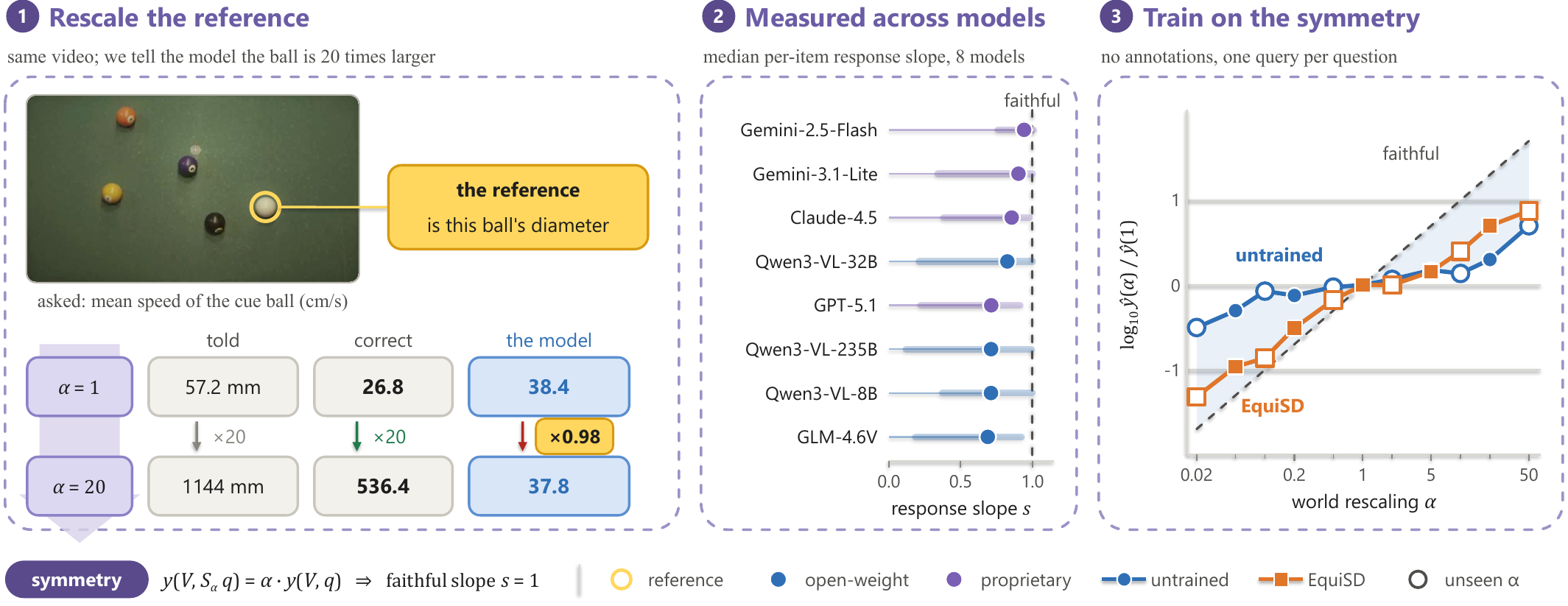}
  \captionsetup{hypcap=false}
  \captionof{figure}{\textbf{How much of the supplied world scale reaches the answer.}
  Multiplying every world-space quantity in the prompt by $\alpha$ keeps the prompt
  consistent with the same video and multiplies the correct answer by exactly $\alpha$,
  so the response slope $s$, the slope of the log answer against $\log\alpha$, is $1$
  for a faithful model. \emph{(1)} One QuantiPhy question at $\alpha{=}1$ and
  $\alpha{=}20$. The reference and the correct answer grow by a factor of $20$, and the
  untrained 3B model's answer changes by a factor of $0.98$. \emph{(2)} Median per-item
  response slope with interquartile range over the $159$ questions for the eight models of
  \cref{tab:diagnosis}.
  \emph{(3)} Median response of the 3B model over the $159$ questions, normalised at
  $\alpha{=}1$, before and after \ours{} training. Filled markers are the five training
  scales and hollow markers six scales not used in training.}
  \captionsetup{hypcap=true}
  \label{fig:teaser}%
\end{strip}

\begin{abstract}
Metric questions about video, such as the speed of a moving object, require a
vision-language model to convert visual measurements into physical units using a
real-world reference supplied in the prompt. We find that current models use this
reference only partially. When every world-space quantity in the prompt is multiplied
by a common factor, the prompt still describes the same video and the correct answer
changes by exactly that factor, but the predictions of eight models change by less, and
their accuracy stays concentrated near the scale that the depicted objects usually have.
Asked the same physics in a scale-free form, the two models we test recover the
closed-form scaling laws on most items, which indicates that the deficit lies in metric
grounding and not in knowledge of the physical mechanism. Because the scaling relation
is exact, it can serve as supervision without metric annotations. Equivariance
Self-Distillation (\ours{}) projects a model's own prediction onto the functions that
satisfy this relation and fine-tunes the model on the resulting targets, with one query
per training question and no ground truth. Trained on synthetic video only, \ours{}
brings a 3B model close to the exact relation on held-out simulated videos, also at
scales not seen in training, and improves its accuracy across scales. Without
adaptation, it also improves accuracy across scales on the QuantiPhy benchmark, where its
gain reaches $93\%$ of that obtained by supervision with exact simulator answers.
\end{abstract}

\section{Introduction}
\label{sec:intro}

Deciding whether an approaching car is too fast or whether a part fits a slot requires an
answer in metres and seconds. Vision-language models (VLMs) describe physical events
fluently, and recent benchmarks ask them for such numbers from
video~\cite{quantiphy2026, vsibench2025}, where the gap to human performance remains wide.

A single camera does not determine the scale of a scene, so a metric question also
supplies a real-world reference, such as the diameter of a ball or the walking speed of a
pedestrian. The reference converts pixel measurements into physical units, and the
correct answer is proportional to it. We ask how much of the supplied reference a model
uses.

We multiply every world-space quantity in the prompt (the reference and any supplied
camera distance) by a common factor $\alpha$. The rewritten prompt still describes the
same video, and the correct answer becomes exactly $\alpha$ times the original. The
answers of eight VLMs change by less (\cref{fig:teaser}): their median response slopes,
the slopes of the log answer against $\log\alpha$, lie between $0.69$ and $0.94$, where a
faithful model has a slope of $1$. Accuracy is therefore highest at the scale that the
depicted objects usually have. A scale-free control indicates that the deficit lies in
metric grounding: asked the same physics as a ratio, which needs no units, the two models
we test reproduce the closed-form laws on most items.

Supervised fine-tuning would need metric ground truth, which real video rarely provides
for arbitrary objects and motions. The rescaling itself provides a label-free training
signal. Let $S_\alpha$ denote the rewriting that multiplies every world-space quantity
supplied in a question $q$ by $\alpha$. The correct answer $y$ to $q$ about a video $V$ is
homogeneous of degree one under it, $y(V,S_\alpha q)=\alpha\,y(V,q)$, which follows
exactly from projective geometry. We call this the homogeneity
relation, and a model that satisfies it scale-equivariant.

Equivariance Self-Distillation (\ours{}) divides the model's own answer at the original
scale by the reference, which projects the answer onto the family of functions that
satisfy the relation, and, without any ground truth, fine-tunes the model to return this
ratio times the rescaled reference at other scales.

Because these targets contain no magnitude that the model did not produce itself, \ours{}
can teach how the answer changes with the reference but not the typical magnitudes of the
training videos, for example that stopping distances are about half a metre. Supervision
with exact simulator answers also carries the simulator's distribution of magnitudes.
With both trained on synthetic video only, \ours{} obtains $66\%$ of the accuracy gain of
simulator supervision on held-out simulated scenes and $93\%$ of it on QuantiPhy.

Our contributions are as follows.
\begin{itemize}[nosep,leftmargin=*]
\item We formulate the homogeneity relation for all supplied world-space quantities as an
  exact, annotation-free constraint on metric physical reasoning, and derive from it the
  response slope, a per-item measure of how much of the supplied scale a model uses,
  computed from its own outputs.
\item We measure the deficit on QuantiPhy and on a new simulated corpus, and separate it
  from knowledge of the physical mechanism with a scale-free control.
\item We introduce \ours{}, which improves scale equivariance with one query per training
  question and no annotations. The improvement holds at world scales not seen in
  training, and a larger share of it transfers to QuantiPhy than for supervision with
  exact simulator answers. Ablations trace the gains in slope and in transfer to
  rescaling the target with the prompt, not to self-distillation as such.
\item We release \ivpsim{} (interventional physics in simulation), 1{,}000 rendered
  rigid-body videos with exact states and $10{,}661$ single-parameter interventions, with
  sensitivity exponents checked against the closed form for the $25$ outcome--parameter
  pairs that have one.
\end{itemize}

\section{Related work}
\label{sec:related}

\paragraph{Metric physical reasoning from video.}
Recent benchmarks ask models for numerical answers about physical events instead of
descriptions. QuantiPhy~\cite{quantiphy2026} supplies a video and one world-space
reference and scores the numerical answer for size, velocity and acceleration; VSI-Bench
and its successors ask for metric spatial quantities in video~\cite{vsibench2025}.
On its 2D questions, whose motion is parallel to the image plane, QuantiPhy also reports
that multiplying the reference by a scalar sharply degrades accuracy, and interprets
this as evidence that models rely on pretrained priors. Starting from the same
observation, we rescale all world-space quantities together, which keeps the rewritten
prompt valid for depth-aware questions as well, and summarise the response by a
per-item slope. We also use the underlying symmetry as a training signal.

\paragraph{Reconstructing an executable world.}
Another line of work converts a video into a simulator and answers by execution.
Code-as-World~\cite{codeasworld2026} discovers executable world representations with an
agentic propose--execute--verify loop; PhysMind~\cite{physmind2026} builds a reusable
executable world for each video without training; LLMPhy~\cite{llmphy2026} and
SIMPACT~\cite{simpact2026} fit simulator parameters iteratively against the
reconstruction error; $\Delta$ynamics~\cite{dynamics2026} predicts a scene configuration
that can be re-simulated. These systems reconstruct the scene before answering. We
address how a model maps a supplied reference to a metric answer, and improve this
mapping by fine-tuning, with no reconstruction at inference time.

\paragraph{Interventions and counterfactuals.}
CoPhy~\cite{cophy2020} and Filtered-CoPhy~\cite{filteredcophy2022} predict trajectories
after an intervention on the initial conditions; CRIPP-VQA~\cite{crippvqa2022} asks
counterfactual questions about hidden mass and friction; ContPhy~\cite{contphy2024}
extends this setting to fluids and deformable objects;
CausalPhys~\cite{causalscaffolding2026} pairs physical questions with causal graphs.
These works evaluate what happens under an intervention, usually as a categorical or
trajectory-level outcome. We use interventions only as a control: the same intervention
is asked once as a scale-free ratio and once in metres, which separates a model's
knowledge of the physical mechanism from its metric grounding.

\paragraph{Units equivariance and consistency.}
Imposing exact units equivariance on a regressor by working in dimensionless variables
is classical in dimensional analysis and has been formalised for machine
learning~\cite{villar2023dimensionless}. That construction builds the symmetry into a
white-box model. We instead measure how far a black-box VLM departs from the symmetry and
use the departure as a training signal. The closest studies examine whether models use
the evidence they are given, through evidence-channel decompositions for monocular size
estimation~\cite{illposed2026} and test-time consistency under semantic
paraphrase~\cite{testtimeconsistency2025}. A consistency objective of this kind only
requires two answers to agree. The scale symmetry fixes the factor by which the answer
must change, so a single model answer determines the target at any other scale. This
also distinguishes \ours{} from standard self-training, in which self-generated targets
are kept or discarded by a confidence rule.

\section{Metric scale as a constraint}
\label{sec:constraint}

\subsection{The homogeneity relation}
\label{sec:homogeneity}

An item of metric physical reasoning consists of a monocular video $V$ and a question $q$
that names a target quantity. Besides its words, $q$ supplies world-space quantities: a
reference value $\rho$ in world units, such as the diameter of a billiard ball or the
walking speed of a pedestrian, and, for depth-aware questions, camera distances
$c_1,\dots,c_m$. The model must return the target in world units. The reference is
necessary because a monocular video determines the scene only up to a global similarity
transform: a world of any size produces the same pixels if the camera translation is
scaled accordingly.

The same ambiguity fixes how the answer depends on the supplied quantities. Suppose that
every length in the world, including the camera translation, is multiplied by
$\alpha>0$ while time is left unchanged. The video stays identical, and every quantity of
dimension $L\,T^{-k}$ is multiplied by $\alpha$. Let $S_\alpha$ denote the corresponding
rewriting of the question, $(\rho,c_i)\mapsto(\alpha\rho,\alpha c_i)$ with its words
unchanged, and let $\bar q = S_{1/\rho}\,q$ be the question in units of its own
reference. Every target in this setting (size, displacement, speed and acceleration) has
dimension $L\,T^{-k}$, and the frame rate fixes the time base, so
\begin{equation}
    y(V, S_\alpha q) \;=\; \alpha\, y(V, q)
    \quad\Longleftrightarrow\quad
    y(V, q) \;=\; \rho\, R(V, \bar q),
    \label{eq:homogeneity}
\end{equation}
where $R$, the answer in units of the reference, does not depend on the world scale. When
$m>0$, rescaling $\rho$ alone is not an instance of $S_\alpha$, and \cref{sec:slope}
measures the effect of this difference.

\Cref{eq:homogeneity} follows from the projective geometry of image formation and is
exact in the setting of this paper, where the time base is fixed by the video and every
supplied quantity and every target has dimension $L\,T^{-k}$. Under $S_\alpha$ the
rewritten scene is the same observation expressed in a different length unit, not the
same dynamics at a different size, so a dimensional constant that the question does not
supply, such as gravity, is rescaled as well. Dimensionless outcomes and durations are
unchanged by $S_\alpha$, so we exclude them from the question set (\cref{sec:ivpsim}).
Two consequences follow. First, the correct answer to the rescaled question is
determined by the answer to the original one, so the relation can be checked and
enforced without annotation. Second, the rescaled question remains a valid description
of the same video, so $\alpha$ sets how far the supplied scale departs from the scale
that the depicted objects usually have.

\subsection{Measuring how much of the given scale a model uses}
\label{sec:slope}

We probe a model by giving it $S_\alpha q$ in place of $q$ and recording its answer
$\hat y(\alpha)$, and call $\alpha{=}1$, where the supplied quantities take their true
values, the \emph{nominal scale}. For each item we fit a straight line to $\log\hat
y(\alpha)$ against $\log\alpha$ over the probed scales and take its slope as the
\emph{response slope},
\begin{equation}
    s \;=\; \frac{\mathrm{d}\log \hat y(\alpha)}{\mathrm{d}\log \alpha},
    \label{eq:slope}
\end{equation}
which is $1$ for a model that follows the supplied reference and $0$ for one whose
answer is set entirely by its prior about the depicted objects; the $R^2$ of the fit
shows how well a single slope describes the response. We also report the
\emph{equivariance error}, the median of
$\lvert \log\hat y(\alpha) - \log\hat y(1) - \log\alpha \rvert$ over items and scales,
which is $0$ for a model that satisfies \cref{eq:homogeneity} exactly and is quoted in
dex, i.e., orders of magnitude. Neither needs an answer key.

All supplied quantities are rescaled together, the only rewriting compatible with
\cref{eq:homogeneity}. Rescaling the reference alone, as in QuantiPhy's counterfactual
probe on its 2D questions, is equivalent only when no camera distance is supplied; on
depth-aware questions it makes the object sizes and camera distances in the prompt
inconsistent, and such a prompt has no correct answer. On the 90 QuantiPhy questions
that supply camera distances, Qwen3-VL-8B has a median slope of $0.14$ (per-item $R^2$
of $0.46$) under the reference-only rewriting and $0.76$ ($0.90$) under the consistent
one (\cref{sec:suppl-probe}).

\subsection{What current models do}
\label{sec:diagnosis}

\begin{table}[t]
  \centering\small
  \setlength{\tabcolsep}{4pt}
  \begin{tabular}{lccc|cc}
    \toprule
    & \multicolumn{3}{c|}{macro MRA at $\alpha$} & \multicolumn{2}{c}{slope $s$} \\
    \cmidrule(lr){2-4}\cmidrule(lr){5-6}
    model & 0.01 & 1 & 100 & median & \% $<0.5$ \\
    \midrule
    \multicolumn{6}{l}{\emph{open-weight}}\\
    GLM-4.6V & 21.7 & 45.4 & 17.8 & 0.69 & 39 \\
    Qwen3-VL-235B & 23.5 & 43.5 & 22.1 & 0.71 & 38 \\
    Qwen3-VL-32B & 17.1 & 40.1 & 18.4 & 0.83 & 33 \\
    Qwen3-VL-8B & 20.6 & 27.3 & 24.7 & 0.71 & 35 \\
    \midrule
    \multicolumn{6}{l}{\emph{proprietary}}\\
    Gemini-3.1-Flash-Lite & 28.9 & 52.6 & 29.8 & 0.90 & 29 \\
    Claude-Sonnet-4.5 & 29.5 & 50.2 & 28.3 & 0.86 & 30 \\
    GPT-5.1 & 20.8 & 42.0 & 22.8 & 0.71 & 38 \\
    Gemini-2.5-Flash & 23.0 & 29.5 & 26.2 & 0.94 & 16 \\
    \bottomrule
  \end{tabular}
  \caption{\textbf{Accuracy under world rescaling, and the response slope that summarises it.} Accuracy at the nominal scale and at the two ends of the sweep, and the response slope of \cref{eq:slope}, which is $1$ for a faithful model. QuantiPhy-validation, 159 questions, macro-averaged over its four categories; rows are ordered by accuracy at the nominal scale. \Cref{tab:diagnosisfull} gives the full $\alpha$ grid.}
  \label{tab:diagnosis}
\end{table}

All eight models respond only partially to the rescaling. We sweep $\alpha$ from $0.01$
to $100$ on the 159 questions of QuantiPhy-validation~\cite{quantiphy2026} for four
open-weight and four proprietary VLMs, with $8$B to $235$B parameters where disclosed.
\Cref{tab:diagnosis} reports the nominal scale and the two ends of the sweep. Accuracy
is mean relative accuracy (MRA)~\cite{vsibench2025}, the fraction of ten relative-error
thresholds a prediction satisfies, macro-averaged over QuantiPhy's four categories, which
cross 2D or 3D motion with a static or a dynamic reference.
For every model, including the strongest, accuracy peaks at the nominal scale and falls
on both sides, and between a sixth and two fifths of the questions are answered with a
slope below $0.5$. A slope of about $0.7$, the median for four of the eight models,
means that a tenfold increase of the supplied scale changes the answer by a factor of
five.

The log-linear fit of \cref{eq:slope} describes the responses well, with a median
per-item $R^2$ between $0.81$ and $0.96$ across models (\cref{tab:diagnosisfull}). Two
patterns recur (\cref{tab:breakdown}). First, the slope is lowest on in-the-wild
internet footage for all eight models and highest, or tied for highest, on clips whose
background has been removed for seven of them (pooled medians $0.42$ and $0.93$).
Second, on questions that supply a speed or an acceleration and ask for a size, six of
the eight models fall below $0.4$ while staying above $0.7$ on the other three pairings
of a static or dynamic reference and target; within these questions, all eight have a
lower slope when the reference is an acceleration. Both patterns are consistent with
competition from pretrained magnitude priors, which weigh more when the objects are easy
to recognise and when the requested quantity is the size of a familiar object.

\paragraph{Scale-free control.}
\begin{table}[t]
  \centering\small
  \setlength{\tabcolsep}{4pt}
  \begin{tabular}{lcc}
    \toprule
    & Gemini-2.5- & Qwen3-VL- \\[-1pt]
    & Flash & 8B \\
    \midrule
    \multicolumn{3}{l}{\emph{non-zero exponent, asked as a ratio}} \\
    \quad items & 160 & 156 \\
    \quad matches closed form \% & 86 & 84 \\
    \midrule
    \multicolumn{3}{l}{\emph{zero exponent, asked as a ratio}} \\
    \quad items & 87 & 87 \\
    \quad unchanged \% & 100 & 100 \\
    \quad changed \% & 0 & 0 \\
    \midrule
    \multicolumn{3}{l}{\emph{zero exponent, asked in metres}} \\
    \quad items & 37 & 31 \\
    \quad unchanged \% & 0 & 13 \\
    \quad changed \% & 95 & 77 \\
    \bottomrule
  \end{tabular}
  \caption{\textbf{The same interventions asked as a ratio and asked in metres.} \ivpsim{} single-parameter interventions. The first block reports the share of non-zero-exponent items on which the reported ratio matches the closed form to $0.05$ in log. The other two take the items whose exponent is exactly $0$ and ask them first as a ratio and then as a value in metres with a reference length as the only metric anchor; the metric form compares each model against its own un-intervened answer, so its measurement error cancels and the correct ratio is exactly $1$. With $r$ the ratio of a model's two answers, ``unchanged'' is $|\ln r|<0.02$ and ``changed'' is $|\ln r|>0.1$. Interventions on gravity and on object size are excluded from the metric form, where withholding the quantities they act on would leave the prompt inconsistent.}
  \label{tab:intervention}
\end{table}

We intervene on one parameter of a simulated \ivpsim{} scene (\cref{sec:ivpsim}), for
which the closed form gives the sensitivity exponent, the power with which the outcome
depends on that parameter. The relative form of the
question states every scene parameter and asks by what factor the outcome changes, which
needs no units and can be answered from the closed form alone. The absolute form
withholds gravity and every speed and asks for the outcome in metres, so the answer must
be measured from the video. For a change in the mass of a sliding block, for example, the
relative form asks by what factor the stopping distance changes (exactly one), and the
absolute form asks for the new distance in metres. In the relative form,
Gemini-2.5-Flash and Qwen3-VL-8B match the closed-form law on
$86\%$ and $84\%$ of the items with a non-zero exponent, and on all $87$ items whose
exponent is zero, such as the flight time under a change of ball size, both answer that
nothing changes
(\cref{tab:intervention}). In the absolute form we compare, on the zero-exponent items,
each model's answer after the intervention with its own answer before it, so that its
measurement error cancels and the correct ratio is exactly one. The answer still changes
by more than ten percent in $95\%$ and $77\%$ of the cases, with a median change of about
a factor of two. The two models thus know the relevant law but do not apply it when the
answer must be a metric quantity.

\paragraph{Two further controls.}
\begin{table}[t]
  \centering\small
  \setlength{\tabcolsep}{4pt}
  \begin{tabular}{lcccc}
    \toprule
    & $n$ & med.\ $|\Delta|$ & same \% & $>2\times$ off \% \\
    \midrule
    \multicolumn{5}{l}{\emph{Gemini-2.5-Flash}} \\
    \quad rewritten in mm & 92 & 0.22 & 13 & 40 \\
    \quad rewritten in km & 101 & 0.19 & 19 & 41 \\
    \midrule
    \multicolumn{5}{l}{\emph{Qwen3-VL-8B}} \\
    \quad rewritten in mm & 98 & 0.43 & 30 & 58 \\
    \quad rewritten in km & 108 & 0.22 & 29 & 48 \\
    \bottomrule
  \end{tabular}
  \caption{\textbf{Writing the same reference in a different length unit changes the answer, although it changes nothing about the scene.} A reference of $1$~cm and one of $10$~mm describe one world, so a faithful model returns one number and $|\Delta| = |\log_{10}(\hat y_{\text{unit}}/\hat y_{\text{original}})|$ is $0$. ``Same'' counts $|\Delta|<0.01$ and ``$>2\times$ off'' counts $|\Delta|>0.3$.}
  \label{tab:invariance}
\end{table}

First, rewriting the reference in another length unit, for example $10$~mm instead of
$1$~cm, leaves the scene unchanged, so a faithful model returns the same value; only
$13\%$ to $30\%$ of the answers stay the same and $40\%$ to $58\%$ change by more than a
factor of two (\cref{tab:invariance}). Second, asked for the scale-free ratio $R$
instead of the metric answer, Qwen3-VL-8B reports a ratio that still varies with the
reference (\cref{sec:suppl-ratio}), so for this model the deficit does not come from the
output format.

\section{Equivariance self-distillation}
\label{sec:method}

\begin{figure*}[t]
  \centering
  \includegraphics[width=\linewidth]{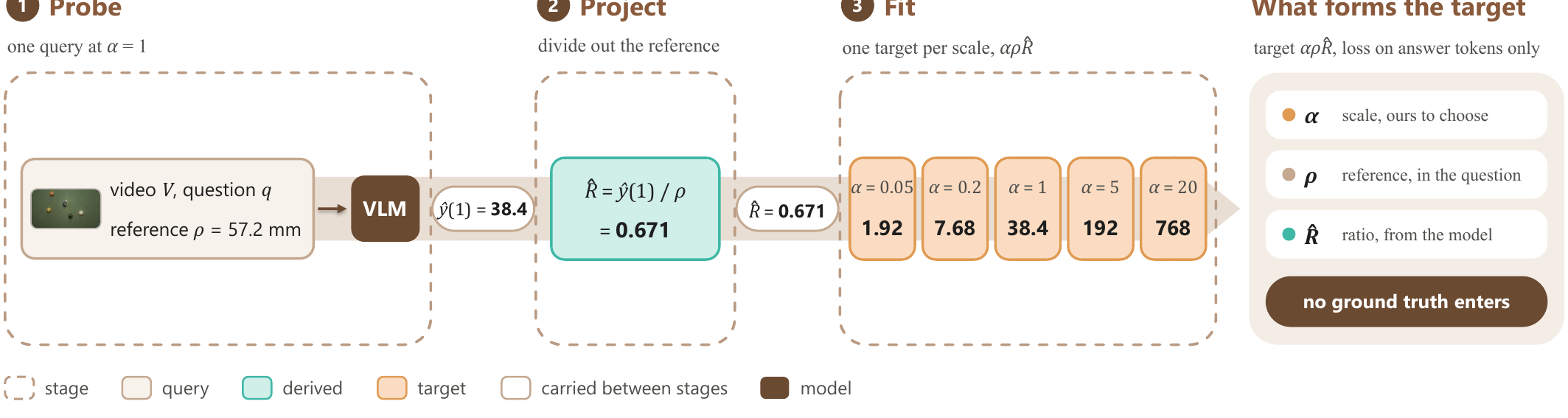}
  \caption{\textbf{\ours{} on one question.} The model is queried once at the nominal
  scale, and its answer divided by the reference gives the scale-free ratio $\hat R$, its
  projection onto the functions that satisfy \cref{eq:homogeneity}. The model is then
  fine-tuned to answer $\alpha\rho\hat R$ on the prompt rewritten by $S_\alpha$, with the
  loss on the answer tokens only. The example is illustrative, as training uses
  \ivpsim{} questions and three scales per question: for the QuantiPhy question of
  \cref{fig:teaser}, the untrained student answers $\hat y(1)=38.4$~cm/s and
  $\alpha\rho\hat R=\alpha\,\hat y(1)$ ranges from $1.92$ to $768$.}
  \label{fig:method}
\end{figure*}

\Cref{eq:homogeneity} constrains how the answer changes with the supplied scale, not its
value at any single scale, so it can be imposed without knowing the correct answer.
\ours{} turns it into a training signal in two steps (\cref{fig:method}): an E-step
projects the model's own responses onto the \emph{equivariant family} of functions that
satisfy \cref{eq:homogeneity}, and an M-step fine-tunes the model on this projection.
Each step is run once.

\paragraph{E-step: project onto the equivariant family.}
A model that satisfies \cref{eq:homogeneity} is determined by its scale-free ratio $R$
alone. We query the untrained model on a probe grid
$\mathcal{A}=\{\alpha_1,\dots,\alpha_K\}$, symmetric in $\log\alpha$ around
$\alpha{=}1$, and take the ratio that best fits all $K$ responses,
\begin{equation}
    \log \hat R \;=\; \operatorname*{median}_{\alpha \in \mathcal{A}}
    \bigl[\log \hat y(\alpha) - \log\alpha - \log\rho \bigr].
    \label{eq:estep}
\end{equation}
The median is the projection onto the equivariant family under an $\ell_1$ criterion in
log space, and it is robust to occasional responses with a wrong format or magnitude.

\paragraph{M-step: fit the projection.}
The projected model answers $\alpha\rho\hat R$ at scale $\alpha$. We fine-tune on these
targets,
\begin{equation}
    \mathcal{L} \;=\; -\!\!\sum_{(V,q)}\ \sum_{\alpha\sim\mathcal{T}}
    \log \pi_\theta\!\bigl(\,\texttt{fmt}(\alpha\rho\hat R)\ \big|\ V, S_\alpha q\,\bigr),
    \label{eq:mstep}
\end{equation}
where $\pi_\theta$ is the trained model, $\mathcal{T}$ is a grid of training scales, and
\texttt{fmt} writes a number as an answer string. The loss covers only the answer
tokens, and the prompt is rewritten by the same $S_\alpha$ as in the probe, so every
camera distance moves with the reference. The loss contains no annotation: $\hat R$
comes from the model, $\rho$ from the question, and $\alpha$ is chosen freely. The
procedure applies to any video for which a plausible reference quantity can be named.

\paragraph{Choosing $K$.}
The responses are close to log-linear in $\alpha$ (\cref{sec:diagnosis}), so the
bracketed term in \cref{eq:estep} is close to linear in $\log\alpha$, its median over a
log-symmetric grid is its value at $\alpha{=}1$, and the estimate reduces to
$\hat R = \hat y(1)/\rho$. The two estimates agree to a median of $0.003$ dex on our
training pool. We use $K{=}1$ by default, which costs one query per training question,
and report the grid form as an ablation.

\paragraph{Implementation.}
The \emph{student} is Qwen2.5-VL-3B-Instruct~\cite{qwen25vl2025} in 4-bit
NF4~\cite{dettmers2023qlora} with LoRA~\cite{hu2022lora} adapters on all attention and
MLP projections, six frames per video, one epoch. Fine-tuning takes $32$ minutes and
$5.85$\,GB of peak memory on a single laptop RTX 4060, after one generation per training
question for the E-step. The remaining hyper-parameters are in \cref{sec:suppl-training}.

\section{Experiments}
\label{sec:experiments}

\subsection{Setup}
\label{sec:setup}

\paragraph{\ivpsim{}.}
\label{sec:ivpsim}
We render 1{,}000 rigid-body videos in MuJoCo~\cite{mujoco2012} across five families
(a block sliding to rest, a projectile, a two-ball collision, a pendulum, and a block on
a ramp), randomising the dynamics, the world scale, the camera and the appearance, with
object states logged at 200\,Hz independently of the rendering. From these we build
3{,}000 QuantiPhy-style questions, restricted to targets of dimension $L\,T^{-k}$, and
10{,}661 single-parameter intervention records. Where a closed form exists, the measured
log-log sensitivity exponents are checked against it, including the cases whose exponent
is exactly zero and on which any predicted change is wrong (\cref{sec:suppl-ivpsim}).

\paragraph{Protocol.}
We evaluate on QuantiPhy-validation (159 questions on 24 videos) and on an \ivpsim{}
split held out by scene family: training uses 400 questions on 134 videos of the
projectile, collision, pendulum and incline families and evaluation 120 questions on 40
sliding-block videos, so the held-out set shares the renderer and the world-scale
distribution with training but not the dynamics. Accuracy
is MRA throughout, macro-averaged over the four categories on QuantiPhy. Model
versions, decoding, prompts and parsing are given in \cref{sec:suppl-protocol}. API
responses are not fully deterministic at temperature zero, so every comparison is made
on paired items within one protocol.

\paragraph{Training.}
The untrained student is queried once on each of the 400 training questions to form the
ratio of \cref{eq:estep}. It gives no valid answer on 62 of them, and is fine-tuned on
the other 338 at three scales per question drawn from $\mathcal{T}=\{0.05,0.2,1,5,20\}$.
This grid is also the main evaluation grid, and \cref{sec:main} adds six scales not used
in training. As a supervised baseline, \emph{simulator supervision} replaces the
student's answer with the simulator's exact answer on all 400 questions and is otherwise
identical, including the scales at which the targets are placed. We train on synthetic
video so that the same questions can also be supervised with exact answers; \ours{}
itself needs no labels (\cref{sec:method}). The optimiser and adapter settings are in
\cref{sec:suppl-training}.

\subsection{Label-free equivariance training}
\label{sec:main}

\begin{figure*}[t]
  \centering
  \includegraphics[width=\linewidth]{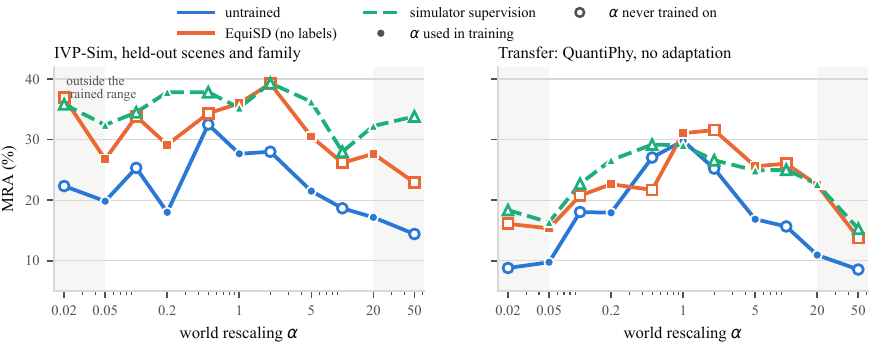}
  \caption{\textbf{Training on the symmetry raises accuracy at scales far from the
  nominal one, including scales not used in training.} MRA of Qwen2.5-VL-3B when every
  world-space quantity in the prompt is multiplied by $\alpha$. Both trained models saw
  only synthetic video; the right panel evaluates the same checkpoints on QuantiPhy
  without adaptation. Filled markers are the five training scales and hollow markers six
  scales not used in training, two of them outside the trained range (shaded).}
  \label{fig:profiles}
\end{figure*}

\begin{table}[t]
  \centering\small
  \setlength{\tabcolsep}{4pt}
  \begin{tabular}{lcccc}
    \toprule
    & mean MRA & $\Delta$ & $s$ & equiv. \\
    \midrule
    \multicolumn{5}{l}{\emph{\ivpsim{}, held out by scene and by dynamics family}}\\
    untrained & 20.8 & -- & 0.66 & 0.30 \\
    \ours{} (no labels) & 30.0 & $+9.2$ & 0.94 & 0.18 \\
    simulator supervision & 34.8 & $+14.0$ & 0.92 & 0.22 \\
    \midrule
    \multicolumn{5}{l}{\emph{Transfer: QuantiPhy, no adaptation}}\\
    untrained & 17.1 & -- & 0.33 & 0.72 \\
    \ours{} (no labels) & 23.4 & $+6.4$ & 0.57 & 0.56 \\
    simulator supervision & 23.9 & $+6.9$ & 0.69 & 0.45 \\
    \bottomrule
  \end{tabular}
  \caption{\textbf{Training on the symmetry without labels obtains most of the gain of
  simulator supervision, and a larger share of it on QuantiPhy.} Qwen2.5-VL-3B trained
  on synthetic video only. Mean MRA is averaged over the $\alpha$ grid and $\Delta$ is
  the paired change against the untrained model; $s$ is the median response slope
  (\cref{eq:slope}, $1$ is faithful); the last column is the equivariance error in dex,
  and $0$ is exact. \Cref{fig:profiles} gives the accuracy at each scale and
  \cref{tab:ablationfull} the $95\%$ bootstrap intervals.}
  \label{tab:main}
\end{table}
\begin{table}[t]
  \centering\small
  \setlength{\tabcolsep}{3pt}
  \begin{tabular}{lccc|cc}
    \toprule
    & \multicolumn{3}{c|}{MRA at $\alpha$} & \multicolumn{2}{c}{slope $s$} \\
    \cmidrule(lr){2-4}\cmidrule(lr){5-6}
    & seen & interp. & extrap. & seen & unseen \\
    \midrule
    \multicolumn{6}{l}{\emph{\ivpsim{}, held out by scene and family, $n{=}120$}}\\
    untrained & 20.8 & 26.1 & 18.4 & 0.66 & 0.69 \\
    \ours{} & 30.0 & 33.4 & 29.9 & 0.94 & 0.96 \\
    labels & 34.8 & 34.9 & 34.8 & 0.92 & 0.92 \\
    \midrule
    \multicolumn{6}{l}{\emph{Transfer: QuantiPhy, $n{=}159$}}\\
    untrained & 17.1 & 21.5 & 8.7 & 0.33 & 0.37 \\
    \ours{} & 23.4 & 25.0 & 14.9 & 0.57 & 0.65 \\
    labels & 23.9 & 25.9 & 16.8 & 0.69 & 0.68 \\
    \bottomrule
  \end{tabular}
  \caption{\textbf{The improvement holds at world scales not used in training, inside and outside the trained range.} Training and \cref{tab:main} share the grid $\{0.05,0.2,1,5,20\}$ (``seen''). The same checkpoints are re-evaluated here on $\{0.1,0.5,2,10\}$, between the trained scales, and on $\{0.02,50\}$, a factor of $2.5$ beyond either end. The groups lie at different distances from the nominal scale, so gains should be compared within a column. Slopes fit \cref{eq:slope} within each grid separately; ``labels'' is simulator supervision.}
  \label{tab:heldout}
\end{table}

\begin{table}[t]
  \centering\small
  \setlength{\tabcolsep}{2.5pt}
  \begin{tabular}{llcccc}
    \toprule
    target & moves with $\alpha$ & mean MRA & $\Delta$ & $s$ & equiv. \\
    \midrule
    -- & -- & 20.8 & -- & 0.66 & 0.30 \\
    self & $\alpha{=}1$ only & 25.9 & $+5.0$ & 0.71 & 0.37 \\
    self & held fixed & 17.0 & $-3.8$ & 0.06 & 0.90 \\
    self (\ours{}) & $\times\alpha$ & 30.0 & $+9.2$ & 0.94 & 0.18 \\
    self, $K{=}5$ & $\times\alpha$ & 28.0 & $+7.1$ & 1.00 & 0.19 \\
    labels & $\alpha{=}1$ only & 18.8 & $-2.0$ & 0.18 & 0.72 \\
    labels & $\times\alpha$ & 34.8 & $+14.0$ & 0.92 & 0.22 \\
    \bottomrule
  \end{tabular}
  \caption{\textbf{The scale a target is placed at determines the model's scale response, whether the target is exact or self-generated.} Held-out \ivpsim{} scenes, $n{=}120$. ``Self'' rows train on the model's own nominal-scale answer $\hat y(1)$ and ``labels'' rows on the simulator's exact answer; within a source, the rows other than $K{=}5$ share the training questions, the nominal-scale targets and the number of updates. ``$\alpha{=}1$ only'' never rescales the prompt, ``held fixed'' rescales it while keeping the target, and ``$\times\alpha$'' rescales both as \cref{eq:homogeneity} requires. $\Delta$ is the paired change against the untrained first row. \Cref{tab:ablationfull} adds the transfer block and intervals.}
  \label{tab:ablation}
\end{table}

\ours{} makes the student follow the supplied scale more closely (\cref{tab:main}). On
held-out simulated video, which also has unseen dynamics, the median response slope
rises from $0.66$ to $0.94$, the equivariance error falls from $0.30$ to $0.18$ dex, and
mean MRA over a 400-fold range of world scale rises by $9.2$ points. \Cref{fig:profiles}
shows the accuracy at each scale: \ours{} is more accurate than the untrained student at
all eleven scales on \ivpsim{}, whereas on QuantiPhy its gains lie far from the nominal
scale and it is $5.3$ points less accurate at $\alpha{=}0.5$.

\paragraph{Scales not seen in training.}
To test whether \ours{} learns a continuous relation or only fits the five training
scales, we evaluate the same checkpoints at six other scales (\cref{tab:heldout}): four
between the trained scales and two outside the trained range, $0.02$ and $50$. The
response slope is similar on the seen and unseen scales, $0.94$ against $0.96$ on the
simulated split and $0.57$ against $0.65$ on QuantiPhy. The three groups lie at
different distances from the nominal scale, so we compare the gain over the untrained
student within each group. Outside the trained range the gain is comparable to that on
the trained scales, $11.54$ against $9.18$ points on \ivpsim{} and $6.26$ against
$6.38$ on QuantiPhy; between the trained scales, where the untrained student is already
more accurate, it is smaller, $7.27$ and $3.52$.

\subsection{Transfer to QuantiPhy}
\label{sec:transfer}

The lower block of \cref{tab:main} evaluates the same checkpoints on QuantiPhy without
any adaptation. The student was trained only on rendered blocks and balls, whereas
QuantiPhy shows cars, people, billiard tables and boats, filmed with real cameras in 67
of its 159 questions and rendered in Blender in the other 92. The improvement carries
over: mean MRA across scales rises by $6.4$ points and the median slope from $0.33$ to
$0.57$, still far below the $0.94$ reached on the simulated split.

\paragraph{Comparison with simulator supervision.}
Simulator supervision, which uses the exact answer for every training question, gains
$14.0$ MRA on the held-out simulated scenes against $9.2$ for \ours{}, and $6.9$ on
QuantiPhy against $6.4$, so \ours{} obtains $66\%$ of the supervised gain on simulated
video and $93\%$ on QuantiPhy. On the 67 questions with real footage the share is $82\%$
($6.15$ against $7.46$ points, \cref{tab:source}).

\paragraph{Gains by video source.}
\begin{figure*}[t]
  \centering
  \includegraphics[width=\linewidth]{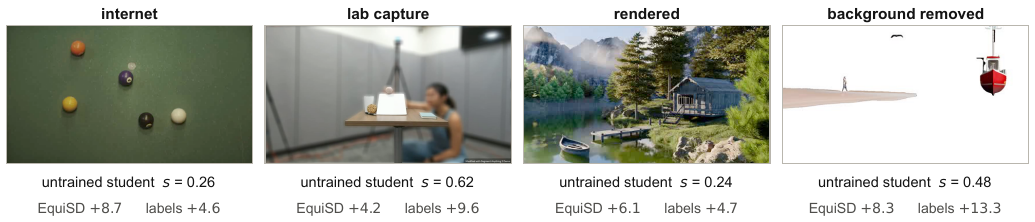}
  \caption{\textbf{\ours{} gains more than simulator supervision (``labels'') on the two
  sources where the untrained student uses the reference least.} One frame per QuantiPhy
  video source with the untrained student's median slope $s$ and the paired MRA gains.
  Internet and lab clips are real footage, the other two Blender renders; \cref{tab:source}
  adds bootstrap intervals.}
  \label{fig:sources}
\end{figure*}

\Cref{fig:sources,tab:source} break the transfer down by video source, along which the
slopes of the eight models vary (\cref{sec:diagnosis}). \ours{} gains more than simulator
supervision on the two sources where the untrained student has the lowest slope and less
on the other two. The interval of the difference excludes zero only for internet
footage, in favour of \ours{}, and for lab captures, in favour of simulator supervision,
and the overall interval includes zero. The pattern is consistent with exact answers
carrying the simulator's distribution of magnitudes, which helps most where the
magnitudes in the test videos are close to the simulator's, whereas the targets of
\ours{} carry no magnitudes.

\subsection{Ablations}
\label{sec:ablation}

\paragraph{The role of target transport.}
\ours{} trains the student on its own answers and transports these answers to other
scales by \cref{eq:homogeneity}. Five variants in \cref{tab:ablation} separate the two:
three use the student's own answer $\hat y(1)$ as the target and two the simulator's
exact answer, and within each group only the rescaling of the prompt and the target
differs. Training on the student's own answers at $\alpha{=}1$ without rescaling the
prompt, which is ordinary self-distillation, gains $5.0$ MRA on \ivpsim{} but leaves the
slope almost unchanged ($0.71$ against $0.66$) and transfers $1.4$ MRA to QuantiPhy,
with an interval that includes zero, against $6.4$ for \ours{}. Rescaling the prompt
while holding the target fixed teaches the student to ignore the reference and drives
the slope to $0.06$. Exact labels show the same dependence: they raise the slope to
$0.92$ when the target is rescaled with the prompt and lower it to $0.18$ when used at
$\alpha{=}1$ only. The gain in equivariance comes from transporting the target with the
prompt, for self-generated and exact targets alike.

\paragraph{Number of probe scales.}
Running the E-step on the five-point grid ($K{=}5$) instead of the nominal scale alone
raises the slope from $0.94$ to $1.00$ and changes mean MRA from $30.0$ to $28.0$, a
paired difference of $-2.1$ whose interval, $[-4.6,+0.4]$, includes zero. The grid form
also keeps 44 of the 62 questions without a valid nominal-scale answer, those with a
valid answer at another scale, so it trains on 382 questions instead of 338. We use
$K{=}1$ because it needs a single query.

\paragraph{Test-time correction of a frozen model.}
The relation needs no labels, so it could also correct a frozen model at inference, by
reading the ratio off the probe at the scale where the fitted response equals the answer
the model gives without a reference (\cref{sec:testtime}). Applied to every question of
\ivpsim{}, this correction lowers accuracy by $4.7$ MRA relative to a single direct
query, because the variance of the estimate grows as $1/s^2$, so it is least reliable on
the questions with a small slope, which are the ones that need correction. Restricting
it to questions whose slope exceeds a threshold reduces the loss to $0.2$ MRA but yields
no gain, so we use the constraint for training.

\section{Limitations}
\label{sec:limitations}

\Cref{eq:homogeneity} holds only in the setting of \cref{sec:homogeneity}: a time base
fixed by the video, and supplied quantities and a target of dimension $L\,T^{-k}$.
Questions whose answer is a duration or a dimensionless ratio are not constrained by it,
and neither are questions whose answer depends on a dimensional constant that the prompt
does not supply.

Equivariance constrains how a model's answers at different scales relate to each other
and does not guarantee that any of them is correct. A model can satisfy
\cref{eq:homogeneity} exactly and be wrong by the same factor at every scale.
Conversely, the variant trained on exact labels at $\alpha{=}1$ only gains $9.4$ MRA at
the nominal scale while its slope falls from $0.66$ to $0.18$ (\cref{tab:ablationfull}),
so we report accuracy and slope together. Applying the constraint at inference alone does
not improve accuracy either (\cref{sec:ablation}).

The diagnosis covers eight models, but \ours{} is evaluated on one 3B student trained on
one synthetic corpus, and the intervention control runs two models. The simulated
evaluation uses one held-out dynamics family, the sliding block.

\section{Conclusion}
\label{sec:conclusion}

The correct answer to a metric question about a monocular video is proportional to the
reference supplied with it. As a measurement tool, this relation shows that eight VLMs
pass only part of the supplied scale to their answers, with median response slopes
between $0.69$ and $0.94$, and that their accuracy is concentrated around the scale that
the depicted objects usually have. As a training signal, it leads to \ours{}, which
fine-tunes a model on its own answers transported to other scales.

\ours{} needs no annotations and takes $32$ minutes on a laptop GPU. A student trained
only on rendered blocks and balls also improves on QuantiPhy and keeps a larger share of
the gain of simulator supervision there than on simulated video, which is consistent with
its targets containing no simulator-specific magnitudes. Physics provides further
label-free constraints of this kind, such as conservation laws and invariance under a
change of frame, which also relate a model's outputs to one another without fixing their
values.

\clearpage
{
    \small
    \bibliographystyle{ieeenat_fullname}
    \bibliography{main}

\begin{thebibliography}{19}
\providecommand{\natexlab}[1]{#1}
\providecommand{\url}[1]{\texttt{#1}}
\expandafter\ifx\csname urlstyle\endcsname\relax
  \providecommand{\doi}[1]{doi: #1}\else
  \providecommand{\doi}{doi: \begingroup \urlstyle{rm}\Url}\fi

\bibitem[Bai et~al.(2025)]{qwen25vl2025}
Shuai Bai et~al.
\newblock {Qwen2.5-VL} technical report.
\newblock \emph{arXiv preprint arXiv:2502.13923}, 2025.

\bibitem[Baradel et~al.(2020)Baradel, Neverova, Mille, Mori, and
  Wolf]{cophy2020}
Fabien Baradel, Natalia Neverova, Julien Mille, Greg Mori, and Christian Wolf.
\newblock {CoPhy}: Counterfactual learning of physical dynamics.
\newblock In \emph{International Conference on Learning Representations
  (ICLR)}, 2020.

\bibitem[Cherian et~al.(2026)Cherian, Corcodel, Jain, and Romeres]{llmphy2026}
Anoop Cherian, Radu Corcodel, Siddarth Jain, and Diego Romeres.
\newblock {LLMPhy}: Parameter-identifiable physical reasoning combining large
  language models and physics engines.
\newblock In \emph{International Conference on Artificial Intelligence and
  Statistics (AISTATS)}, 2026.
\newblock arXiv:2411.08027.

\bibitem[Chou et~al.(2025)Chou, Chandhok, Little, and
  Sigal]{testtimeconsistency2025}
Shih-Han Chou, Shivam Chandhok, James~J. Little, and Leonid Sigal.
\newblock Test-time consistency in vision language models, 2025.
\newblock arXiv:2506.22395.

\bibitem[Dettmers et~al.(2023)Dettmers, Pagnoni, Holtzman, and
  Zettlemoyer]{dettmers2023qlora}
Tim Dettmers, Artidoro Pagnoni, Ari Holtzman, and Luke Zettlemoyer.
\newblock {QLoRA}: Efficient finetuning of quantized {LLMs}.
\newblock In \emph{Advances in Neural Information Processing Systems
  (NeurIPS)}, pages 10088--10115, 2023.

\bibitem[Hu et~al.(2022)Hu, Shen, Wallis, Allen-Zhu, Li, Wang, Wang, and
  Chen]{hu2022lora}
Edward~J. Hu, Yelong Shen, Phillip Wallis, Zeyuan Allen-Zhu, Yuanzhi Li, Shean
  Wang, Lu Wang, and Weizhu Chen.
\newblock {LoRA}: Low-rank adaptation of large language models.
\newblock In \emph{International Conference on Learning Representations
  (ICLR)}, 2022.

\bibitem[Janny et~al.(2022)Janny, Baradel, Neverova, Nadri, Mori, and
  Wolf]{filteredcophy2022}
Steeven Janny, Fabien Baradel, Natalia Neverova, Madiha Nadri, Greg Mori, and
  Christian Wolf.
\newblock Filtered-{CoPhy}: Unsupervised learning of counterfactual physics in
  pixel space.
\newblock In \emph{International Conference on Learning Representations
  (ICLR)}, 2022.
\newblock arXiv:2202.00368.

\bibitem[Kao et~al.(2026)Kao, Huynh, Wang, Vesdapunt, Stojanov, Hariharan,
  Obiednikov, and Zhou]{dynamics2026}
Chia-Hsiang Kao, Cong~Phuoc Huynh, Chien-Yi Wang, Noranart Vesdapunt, Stefan
  Stojanov, Bharath Hariharan, Oleksandr Obiednikov, and Ning Zhou.
\newblock {$\Delta$}ynamics: Language-based representation for inferring
  rigid-body dynamics from videos.
\newblock In \emph{Proceedings of the IEEE/CVF Conference on Computer Vision
  and Pattern Recognition (CVPR)}, pages 42364--42374, 2026.
\newblock arXiv:2605.20576.

\bibitem[Li et~al.(2026)Li, Xiang, Mao, Wei, Chen, Masood, Fei-Fei, and
  Adeli]{quantiphy2026}
Puyin Li, Tiange Xiang, Ella Mao, Shirley Wei, Xinye Chen, Adnan Masood, Li
  Fei-Fei, and Ehsan Adeli.
\newblock {QuantiPhy}: A quantitative benchmark evaluating physical reasoning
  abilities of vision-language models.
\newblock In \emph{Proceedings of the IEEE/CVF Conference on Computer Vision
  and Pattern Recognition (CVPR)}, pages 33174--33184, 2026.
\newblock arXiv:2512.19526.

\bibitem[Liu et~al.(2026)Liu, Yao, Chen, Gao, Mao, Huang, and Du]{simpact2026}
Haowen Liu, Shaoxiong Yao, Haonan Chen, Jiawei Gao, Jiayuan Mao, Jia-Bin Huang,
  and Yilun Du.
\newblock {SIMPACT}: Simulation-enabled action planning using vision-language
  models.
\newblock In \emph{Proceedings of the IEEE/CVF Conference on Computer Vision
  and Pattern Recognition (CVPR)}, pages 20790--20801, 2026.
\newblock arXiv:2512.05955.

\bibitem[Meivar et~al.(2026)Meivar, Perek, Shvartzman, Schwartz, and
  Avidan]{illposed2026}
Boaz Meivar, Shaked Perek, Shani Shvartzman, Eli Schwartz, and Shai Avidan.
\newblock Ill-posed by design: Probing evidence use in {VLMs}, 2026.
\newblock arXiv:2606.24335.

\bibitem[Patel et~al.(2022)Patel, Gokhale, Baral, and Yang]{crippvqa2022}
Maitreya Patel, Tejas Gokhale, Chitta Baral, and Yezhou Yang.
\newblock {CRIPP-VQA}: Counterfactual reasoning about implicit physical
  properties via video question answering.
\newblock In \emph{Proceedings of the 2022 Conference on Empirical Methods in
  Natural Language Processing (EMNLP)}, pages 9856--9870, 2022.
\newblock arXiv:2211.03779.

\bibitem[Tang et~al.(2026)Tang, Lin, Feng, Ma, Ong, Tsang, and
  Yin]{causalscaffolding2026}
Tianyi Tang, Zhuoyi Lin, Zeyu Feng, Tianyi Ma, Yew-Soon Ong, Ivor Tsang, and
  Haiyan Yin.
\newblock Causal scaffolding for physical reasoning: A benchmark for
  causally-informed physical world understanding in {VLMs}.
\newblock In \emph{Proceedings of the 32nd ACM SIGKDD Conference on Knowledge
  Discovery and Data Mining (KDD)}, 2026.
\newblock arXiv:2606.05966.

\bibitem[Todorov et~al.(2012)Todorov, Erez, and Tassa]{mujoco2012}
Emanuel Todorov, Tom Erez, and Yuval Tassa.
\newblock {MuJoCo}: A physics engine for model-based control.
\newblock In \emph{IEEE/RSJ International Conference on Intelligent Robots and
  Systems (IROS)}, pages 5026--5033, 2012.

\bibitem[Villar et~al.(2023)Villar, Yao, Hogg, Blum-Smith, and
  Dumitrascu]{villar2023dimensionless}
Soledad Villar, Weichi Yao, David~W. Hogg, Ben Blum-Smith, and Bianca
  Dumitrascu.
\newblock Dimensionless machine learning: Imposing exact units equivariance.
\newblock \emph{Journal of Machine Learning Research}, 24\penalty0
  (109):\penalty0 1--32, 2023.
\newblock arXiv:2204.00887.

\bibitem[Wang et~al.(2026)Wang, Cai, Chen, Chi, Sun, Dai, Hung, Guo, Ren, Yao,
  Liu, Long, Duan, Gao, Lyu, Liu, and Wu]{codeasworld2026}
Hanyang Wang, Yimo Cai, Weiliang Chen, Jiawei Chi, Haowen Sun, Qiyu Dai,
  Yi-Hsin Hung, Xingzhuo Guo, Jinshan Ren, Runmao Yao, Ziwei Liu, Mingsheng
  Long, Yueqi Duan, Jun Gao, Jiangran Lyu, Fangfu Liu, and Jialong Wu.
\newblock Code as worlds: Agentic discovery of executable world representations
  for physical reasoning.
\newblock \emph{arXiv preprint arXiv:2608.27549}, 2026.

\bibitem[Yang et~al.(2026)Yang, Zeng, Zhao, Xu, He, Li, Deng, Fan, and
  Wang]{physmind2026}
Chen Yang, Shenxiang Zeng, Haoyang Zhao, Zhouyuan Xu, Youquan He, Haoyu Li,
  Mingyi Deng, Jiansheng Fan, and Chen Wang.
\newblock {PhysMind}: From video to executable worlds for training-free
  physical reasoning, 2026.
\newblock arXiv:2608.04575.

\bibitem[Yang et~al.(2025)Yang, Yang, Gupta, Han, Fei-Fei, and
  Xie]{vsibench2025}
Jihan Yang, Shusheng Yang, Anjali~W. Gupta, Rilyn Han, Li Fei-Fei, and Saining
  Xie.
\newblock Thinking in space: How multimodal large language models see,
  remember, and recall spaces.
\newblock In \emph{Proceedings of the IEEE/CVF Conference on Computer Vision
  and Pattern Recognition (CVPR)}, pages 10632--10643, 2025.

\bibitem[Zheng et~al.(2024)Zheng, Yan, Chen, Wang, Lim, Tenenbaum, and
  Gan]{contphy2024}
Zhicheng Zheng, Xin Yan, Zhenfang Chen, Jingzhou Wang, Qin Zhi~Eddie Lim,
  Joshua~B. Tenenbaum, and Chuang Gan.
\newblock {ContPhy}: Continuum physical concept learning and reasoning from
  videos.
\newblock In \emph{International Conference on Machine Learning (ICML)}, pages
  61526--61558, 2024.
\newblock arXiv:2402.06119.

\end{thebibliography}
}

\clearpage
\appendix
\section*{Appendix}
\label{sec:suppl}


\section{Full result tables}
\label{sec:suppl-tables}

\begin{table*}[t]
  \centering\footnotesize
  \setlength{\tabcolsep}{5pt}
  \begin{tabular}{lcccccccc|ccc}
    \toprule
    & & \multicolumn{7}{c|}{macro MRA at world rescaling $\alpha$} & \multicolumn{3}{c}{response slope $s$} \\
    \cmidrule(lr){3-9}\cmidrule(lr){10-12}
    model & size & 0.01 & 0.1 & 0.3 & 1 & 3 & 10 & 100 & median & \% $<0.5$ & fit $R^2$ \\
    \midrule
    \multicolumn{12}{l}{\emph{open-weight}}\\
    GLM-4.6V & -- & 21.7 & 22.8 & 26.7 & 45.4 & 34.1 & 25.1 & 17.8 & 0.69 & 39 & 0.86 \\
    Qwen3-VL-235B & 235B-A22B & 23.5 & 23.8 & 25.3 & 43.5 & 33.7 & 24.9 & 22.1 & 0.71 & 38 & 0.82 \\
    Qwen3-VL-32B & 32B & 17.1 & 22.4 & 24.5 & 40.1 & 29.9 & 25.7 & 18.4 & 0.83 & 33 & 0.91 \\
    Qwen3-VL-8B & 8B & 20.6 & 21.0 & 23.6 & 27.3 & 26.9 & 22.4 & 24.7 & 0.71 & 35 & 0.81 \\
    \midrule
    \multicolumn{12}{l}{\emph{proprietary}}\\
    Gemini-3.1-Flash-Lite & -- & 28.9 & 30.3 & 32.2 & 52.6 & 37.1 & 31.1 & 29.8 & 0.90 & 29 & 0.95 \\
    Claude-Sonnet-4.5 & -- & 29.5 & 25.8 & 23.4 & 50.2 & 39.5 & 31.4 & 28.3 & 0.86 & 30 & 0.92 \\
    GPT-5.1 & -- & 20.8 & 25.1 & 26.9 & 42.0 & 33.9 & 29.4 & 22.8 & 0.71 & 38 & 0.84 \\
    Gemini-2.5-Flash & -- & 23.0 & 27.4 & 28.0 & 29.5 & 27.1 & 23.3 & 26.2 & 0.94 & 16 & 0.96 \\
    \bottomrule
  \end{tabular}
  \caption{\textbf{\Cref{tab:diagnosis} over the full sweep.} Parameter counts are given where disclosed. ``Fit $R^2$'' is the median per-item quality of the straight-line fit in $\log\alpha$ that defines $s$.}
  \label{tab:diagnosisfull}
\end{table*}

\begin{table*}[t]
  \centering\footnotesize
  \setlength{\tabcolsep}{5pt}
  \begin{tabular}{llccccc}
    \toprule
    target & moves with $\alpha$ & mean MRA & $\Delta$ & MRA at $\alpha{=}1$ & $s$ & equiv. \\
    \midrule
    \multicolumn{7}{l}{\emph{Held-out \ivpsim{}, $n{=}120$}}\\
    -- & -- & 20.8 & -- & 27.7 & 0.66 & 0.30 \\
    self & $\alpha{=}1$ only & 25.9 & $+5.0$\,{\scriptsize$[+2.4,+7.7]$} & 29.3 & 0.71 & 0.37 \\
    self & held fixed & 17.0 & $-3.8$\,{\scriptsize$[-7.3,-0.5]$} & 30.8 & 0.06 & 0.90 \\
    self (\ours{}) & $\times\alpha$ & 30.0 & $+9.2$\,{\scriptsize$[+6.6,+11.8]$} & 36.0 & 0.94 & 0.18 \\
    self, $K{=}5$ & $\times\alpha$ & 28.0 & $+7.1$\,{\scriptsize$[+4.2,+10.2]$} & 28.6 & 1.00 & 0.19 \\
    labels & $\alpha{=}1$ only & 18.8 & $-2.0$\,{\scriptsize$[-5.2,+1.1]$} & 37.1 & 0.18 & 0.72 \\
    labels & $\times\alpha$ & 34.8 & $+14.0$\,{\scriptsize$[+9.8,+18.1]$} & 35.2 & 0.92 & 0.22 \\
    \midrule
    \multicolumn{7}{l}{\emph{Transfer to QuantiPhy, $n{=}159$}}\\
    -- & -- & 17.1 & -- & 29.8 & 0.33 & 0.72 \\
    self & $\alpha{=}1$ only & 18.5 & $+1.4$\,{\scriptsize$[-0.4,+3.3]$} & 28.6 & 0.29 & 0.70 \\
    self & held fixed & 15.0 & $-2.1$\,{\scriptsize$[-4.3,-0.0]$} & 29.8 & 0.01 & 0.98 \\
    self (\ours{}) & $\times\alpha$ & 23.4 & $+6.4$\,{\scriptsize$[+4.6,+8.2]$} & 31.1 & 0.57 & 0.56 \\
    self, $K{=}5$ & $\times\alpha$ & 22.6 & $+5.5$\,{\scriptsize$[+3.8,+7.3]$} & 33.1 & 0.52 & 0.58 \\
    labels & $\alpha{=}1$ only & 16.0 & $-1.1$\,{\scriptsize$[-3.2,+0.9]$} & 32.6 & 0.06 & 0.82 \\
    labels & $\times\alpha$ & 23.9 & $+6.9$\,{\scriptsize$[+4.5,+9.1]$} & 29.1 & 0.69 & 0.45 \\
    \bottomrule
  \end{tabular}
  \caption{\textbf{\Cref{tab:ablation} with the transfer block, bootstrap intervals and the nominal scale.} Columns are as in \cref{tab:ablation}; $\Delta$ carries a $95\%$ bootstrap interval over items, and ``MRA at $\alpha{=}1$'' is the accuracy at the nominal scale alone.}
  \label{tab:ablationfull}
\end{table*}

\begin{table*}[t]
  \centering\footnotesize
  \setlength{\tabcolsep}{4pt}
  \begin{tabular}{lcccc|cccc|cc}
    \toprule
    & \multicolumn{4}{c|}{video source} & \multicolumn{4}{c|}{reference $\to$ target} & \multicolumn{2}{c}{M$\to$S by reference} \\
    \cmidrule(lr){2-5}\cmidrule(lr){6-9}\cmidrule(lr){10-11}
    model & internet & \shortstack{lab\\capture} & rendered & \shortstack{background\\removed} & S$\to$S & S$\to$M & M$\to$M & M$\to$S & speed & accel. \\
    \midrule
    \multicolumn{11}{l}{\emph{open-weight}}\\
    GLM-4.6V & 0.13 & 0.73 & 0.64 & 0.85 & 0.74 & 0.73 & 1.00 & 0.26 & 0.53 & 0.09 \\
    Qwen3-VL-235B & 0.16 & 0.85 & 0.74 & 0.81 & 0.80 & 0.84 & 1.00 & 0.19 & 0.47 & 0.01 \\
    Qwen3-VL-32B & 0.21 & 0.85 & 0.91 & 0.92 & 0.85 & 0.96 & 1.00 & 0.30 & 0.53 & 0.21 \\
    Qwen3-VL-8B & 0.43 & 0.67 & 0.81 & 0.83 & 0.77 & 0.51 & 0.99 & 0.69 & 0.78 & 0.55 \\
    \midrule
    \multicolumn{11}{l}{\emph{proprietary}}\\
    Gemini-3.1-Flash-Lite & 0.54 & 0.95 & 0.87 & 1.00 & 0.96 & 0.97 & 1.00 & 0.31 & 0.60 & 0.02 \\
    Claude-Sonnet-4.5 & 0.56 & 0.90 & 0.85 & 0.99 & 0.97 & 0.98 & 0.90 & 0.38 & 0.66 & 0.09 \\
    GPT-5.1 & 0.43 & 0.82 & 0.71 & 0.88 & 0.88 & 0.90 & 0.90 & 0.28 & 0.49 & 0.08 \\
    Gemini-2.5-Flash & 0.74 & 0.98 & 0.95 & 0.98 & 0.98 & 0.95 & 0.98 & 0.87 & 0.89 & 0.33 \\
    \midrule
    pooled & 0.42 & 0.89 & 0.82 & 0.93 & 0.91 & 0.90 & 1.00 & 0.42 & 0.64 & 0.09 \\
    questions & 29 & 38 & 74 & 18 & 34 & 41 & 18 & 66 & 41 & 25 \\
    \bottomrule
  \end{tabular}
  \caption{\textbf{Response slope by video source and by type of reference and target.} Median per-item slope $s$ on QuantiPhy-validation ($1$ is faithful). Internet and lab clips are real footage; rendered and background-removed clips are Blender renders. S is a size and M a speed or an acceleration; the last two columns split M$\to$S by the supplied quantity. ``Pooled'' is the median over all model--item fits of the eight models, and the last row gives the number of questions.}
  \label{tab:breakdown}
\end{table*}

\begin{table}[t]
  \centering\footnotesize
  \setlength{\tabcolsep}{5pt}
  \begin{tabular}{lcccc}
    \toprule
    family & pairs & of which zero & max.\ deviation & scenes \\
    \midrule
    slide & 10 & 4 & $0.189$ & 116 \\
    projectile & 12 & 6 & $0.003$ & 119 \\
    pendulum & 3 & 1 & $0.005$ & 179 \\
    \bottomrule
  \end{tabular}
  \caption{\textbf{\ivpsim{}'s ground truth reproduces the closed-form scaling laws.} For every outcome-parameter pair with a closed form we measure $\mathrm{d}\log(\text{outcome})/\mathrm{d}\log(\text{parameter})$ from symmetric multiplicative perturbations and compare it with the exact exponent; the table reports the largest deviation in each family. The pairs whose exact exponent is zero are the sharpest test, since any predicted change there is provably wrong. Contact-rich collisions have no closed form and use the measured value as ground truth.}
  \label{tab:sensitivity}
\end{table}

\begin{table}[t]
  \centering\footnotesize
  \setlength{\tabcolsep}{1.6pt}
  \begin{tabular}{@{}lcc|ccc@{}}
    \toprule
    & & untrained & \multicolumn{3}{c}{gain in MRA} \\
    \cmidrule(lr){4-6}
    video source & $n$ & $s$ & \ours{} & labels & difference \\
    \midrule
    internet & 29 & 0.26 & $+8.7$ & $+4.6$ & $+4.1$\,{\scriptsize$[+0.8,+8.1]$} \\
    lab capture & 38 & 0.62 & $+4.2$ & $+9.6$ & $-5.4$\,{\scriptsize$[-9.3,-1.5]$} \\
    rendered & 74 & 0.24 & $+6.1$ & $+4.7$ & $+1.4$\,{\scriptsize$[-1.9,+4.7]$} \\
    background removed & 18 & 0.48 & $+8.3$ & $+13.3$ & $-5.0$\,{\scriptsize$[-11.0,+1.1]$} \\
    \midrule
    real footage & 67 & 0.43 & $+6.1$ & $+7.5$ & $-1.3$\,{\scriptsize$[-4.4,+1.6]$} \\
    all & 159 & 0.33 & $+6.4$ & $+6.9$ & $-0.5$\,{\scriptsize$[-2.6,+1.7]$} \\
    \bottomrule
  \end{tabular}
  \caption{\textbf{\ours{} gains more than simulator supervision on the sources where the untrained student uses the reference least, and less on the other two.} QuantiPhy transfer by video source. Internet and lab clips are real footage; rendered and background-removed clips are Blender renders. ``Untrained $s$'' is the median response slope of the untrained student on those items; the gain columns are paired changes in mean MRA over the $\alpha$ grid, and the last column is their per-item difference with a $95\%$ bootstrap interval. Positive means that \ours{} gains more; ``labels'' is simulator supervision.}
  \label{tab:source}
\end{table}

\begin{figure*}[t]
  \centering
  \includegraphics[width=\linewidth]{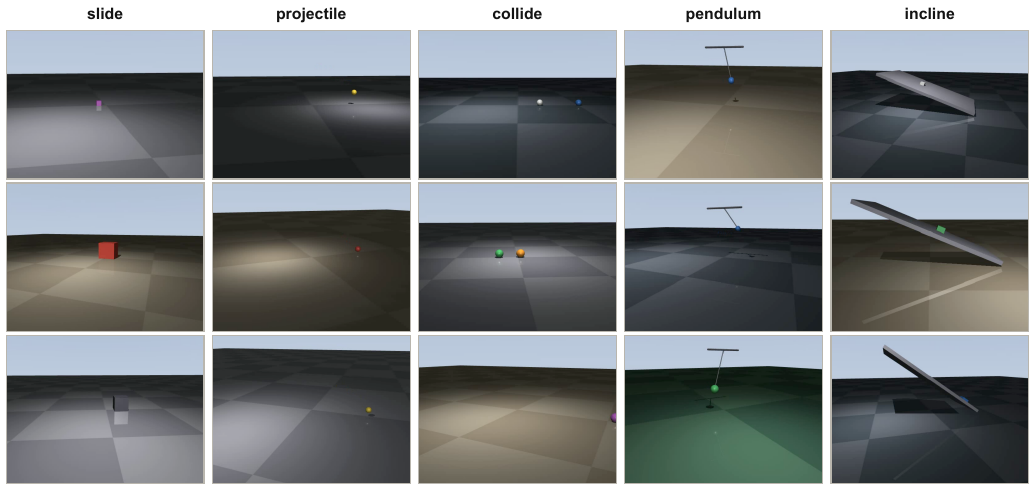}
  \caption{\textbf{\ivpsim{}.} Three frames from each of the five scene families of the
  released corpus. Every scene randomises the parameters that govern its dynamics, the
  overall world scale over $10^{\pm 0.5}$, the camera pose and field of view, the
  lighting position, the floor texture and colour, and the object colour.}
  \label{fig:ivpsim}
\end{figure*}

\Cref{tab:breakdown} breaks the slopes of \cref{sec:diagnosis} down by video source and
by type of reference and target.
\Cref{tab:diagnosisfull,tab:ablationfull} extend \cref{tab:diagnosis,tab:ablation} to
the full $\alpha$ sweep and to the QuantiPhy transfer block with bootstrap intervals; the
shares in \cref{sec:transfer} are computed from the $\Delta$ column of
\cref{tab:ablationfull} and, for real footage, from \cref{tab:source}.

\section{Protocol}
\label{sec:suppl-protocol}

\subsection{Models, decoding and statistics}
\label{sec:suppl-models}

The eight models are queried through OpenRouter on 30 August 2026 as
\texttt{qwen/qwen3-vl-8b-instruct}, \texttt{qwen/qwen3-vl-32b-instruct},
\texttt{qwen/qwen3-vl-235b-a22b-instruct}, \texttt{z-ai/glm-4.6v},
\texttt{google/gemini-2.5-flash}, \texttt{google/gemini-3.1-flash-lite},
\texttt{openai/gpt-5.1} and \texttt{anthropic/claude-sonnet-4.5}, at temperature $0$
with a budget of $8192$ tokens and $16$ uniformly spaced frames per video. The 3B student
runs locally on six frames at a longest side of $280$ pixels, so its accuracy should be
compared across our own training variants and not with the API models.

The response slope of an item is the ordinary least-squares coefficient of
$\log_{10}\hat y$ on $\log_{10}\alpha$ over the probe grid, fitted on the answers that
parse to a positive number and requiring at least two distinct scales; the $R^2$ quoted
beside it comes from the same per-item fit. A paired comparison resamples items
$6{,}000$ times with replacement and reports the $2.5$th and $97.5$th percentiles of the
mean per-item difference.

\subsection{Prompts}
\label{sec:prompts}

Every query uses the same system message: \emph{``You are a careful quantitative
physical reasoner. You measure objects and motion in videos and convert them to
real-world units using the reference information given to you. Always trust the
reference values you are given, even when they differ from typical real-world
magnitudes.''} The last sentence makes the rescaling probe a test of faithfulness to the
reference and not of plausibility judgement: a model that lowers its answer because a
$1.1$~m billiard ball seems unlikely acts against the instruction. Three user prompts
are used.

\paragraph{Direct} is the QuantiPhy protocol. It gives the frame count and duration, the
reference in world units, the depth context where the benchmark supplies one, the
question, and the instruction to end with a line reading
\texttt{Final answer: <number> <unit>}.

\paragraph{No reference} removes the numeric reference and states that none is available,
asking for the model's best estimate, which is the magnitude the model would produce from
the video and its own knowledge.

\paragraph{Ratio} asks for the answer divided by the reference's numerical value, naming
both units explicitly so the requested quantity is unambiguous, and states that the ratio
is fixed by the video alone (\cref{sec:suppl-ratio}).

\subsection{Reading a response}
\label{sec:parsing}

A response is parsed by taking the text after the last \texttt{Final answer:} marker,
falling back to the last line and then to the last number in the reply, and reading the
first scalar there. Scientific notation and thousands separators are handled and the sign
is discarded, following QuantiPhy. A response whose \texttt{finish\_reason} is
\texttt{length} was truncated before the final answer and is scored as a failure, since
any number in it is an intermediate quantity.

\section{Controls behind the diagnosis}
\label{sec:suppl-controls}

\subsection{What the probe rescales}
\label{sec:suppl-probe}

The protocol choice of \cref{sec:slope} is measured for Qwen3-VL-8B on the $90$
QuantiPhy questions that supply a camera distance as well as a reference, with
$\alpha\in\{0.01,0.1,0.3,1,3,10,100\}$. The median slope is $0.14$ with a per-item $R^2$
of $0.46$ under the reference-only rewriting, which makes the prompt inconsistent,
against $0.76$ and $0.90$ under the consistent one. Macro MRA over the two 3D categories
falls from
$27.6$ at $\alpha{=}1$ to $11.5$ at $\alpha{=}100$ under the reference-only rewriting
and averages $16.9$ over the grid, against $25.9$ under the consistent one.

\subsection{Asking for the ratio instead}
\label{sec:suppl-ratio}

Since the video determines $R$ on its own, an obvious remedy is to ask for $R$ and
multiply it by the reference afterwards. For Qwen3-VL-8B at $\alpha{=}1$ this gives a
macro MRA of $14.0$, against $27.3$ for the direct answer (\cref{tab:diagnosis}). The
reported ratio still changes with the reference, by a median of $+0.12$ dex at
$\alpha{=}0.1$ and $-0.90$ dex at $\alpha{=}10$ relative to $\alpha{=}1$; a correct
ratio would not change, and a fixed metric answer would change by $+1$ and $-1$ dex.

\section{Correcting a frozen model at test time}
\label{sec:testtime}

For the test-time correction of \cref{sec:ablation}, we write the response over the
probe grid as $\hat y(\alpha)=c\,\alpha^{s}$. The scale at which this fitted response
equals the answer the model gives without a reference is where the supplied magnitude
agrees with the model's prior, and the ratio is read there. The estimate uses only the
model's own answers and is exact when the response is log-linear. Its variance grows as
$1/s^{2}$. Correcting every question on \ivpsim{} lowers accuracy by $4.7$ MRA
relative to a single direct query. Correcting only the items whose measured slope
exceeds a threshold, chosen on one half of the scenes and evaluated on the other half,
leaves the corrected model $0.2$ MRA below a direct query, and no threshold in the sweep
makes it better.

\section{\ivpsim{}}
\label{sec:suppl-ivpsim}

Each of the five families in \cref{fig:ivpsim} randomises the parameters that govern its
dynamics, the camera pose and field of view, the lighting position, the floor texture and
repeat count, the object colour, and the overall world scale over $10^{\pm 0.5}$, so
that the corpus is not concentrated at one size. Scenes render at $640\times480$,
$25$~fps, for $1.6$ to $3.0$~s. Each intervention record multiplies one parameter of a
scene by a factor between $1.4$ and $2.5$ or its inverse, simulates the changed scene
without rendering it, and stores the outcomes before and after the change.

Outcomes such as a stopping distance or a period are computed from the $200$~Hz state
log and not from the rendered frames: at $25$~fps a pendulum period would be quantised
to $40$~ms, about $3\%$ of a typical period, which is enough to corrupt a sensitivity
exponent measured by finite differences. \Cref{tab:sensitivity} compares the measured
exponents with their closed forms. Of the $25$ outcome--parameter pairs that have a
closed form, $23$ agree with it to within $0.01$. The two exceptions are the stopping
time of a sliding block under a change of friction and of gravity, where the spread
across scenes ($0.44$ and $0.04$) is at least as large as the deviation, because the
instant at which a block counts as stopped varies between scenes.

One outcome is excluded. The acceleration of a block down the ramp, estimated by fitting
its speed over the descending segment, has a median of $0.71$ of the closed-form value
$g(\sin\theta-\mu\cos\theta)$ over 200 scenes, most likely because the block settles
before it slides. We cannot verify it, so it is used neither in questions nor as ground
truth.

Outcomes whose dimension is not $L\,T^{-k}$ are excluded from the questions but kept in
the intervention records. We also remove degenerate items: a speed target below $8\%$ of
the peak speed of the scene, a length below one centimetre, a net displacement smaller
than the object itself, and any item whose reference and target are the same physical
quantity to within $2\%$.

\section{Training details}
\label{sec:suppl-training}

We train LoRA adapters (rank $16$, $\alpha_{\text{LoRA}}=32$, dropout $0.05$) on all
attention and MLP projections of Qwen2.5-VL-3B-Instruct in 4-bit NF4 with double
quantisation, using AdamW at $10^{-4}$ on a one-cycle schedule, weight decay $0.01$,
gradient clipping at $1.0$, batch size $1$ with $16$ accumulation steps, one epoch and
gradient checkpointing. Six frames per video at a maximum of $100{,}352$ pixels give
sequences of about $650$ tokens and a peak memory of $5.85$~GB, so the run fits on an
8~GB laptop GPU. The function \texttt{fmt} of \cref{eq:mstep} prints a target with at
most six decimal places and no trailing zeros, so that the loss falls on the digits that
carry the magnitude.

\end{document}